\documentclass{vgtc}                          % final (conference style)
\ifpdf%                                % if we use pdflatex
  \pdfoutput=1\relax                   % create PDFs from pdfLaTeX
  \usepackage{graphicx}                % allow us to embed graphics files
  \DeclareGraphicsExtensions{.pdf,.png,.jpg,.jpeg} % for pdflatex we expect .pdf, .png, or .jpg files
\else%                                 % else we use pure latex
  \usepackage{graphicx}                % allow us to embed graphics files
  \DeclareGraphicsExtensions{.eps}     % for pure latex we expect eps files
\fi%

\graphicspath{{figures/}{pictures/}{images/}{./}} % where to search for the images

\usepackage{microtype}                 % use micro-typography (slightly more compact, better to read)
\PassOptionsToPackage{warn}{textcomp}  % to address font issues with \textrightarrow
\usepackage{textcomp}                  % use better special symbols
\usepackage{mathptmx}                  % use matching math font
\usepackage{times}                     % we use Times as the main font
\usepackage{cite}                      % needed to automatically sort the references
\usepackage{tabu}                      % only used for the table example
\usepackage{booktabs}                  % only used for the table example
\usepackage{mathptmx}                  % use matching math font
\usepackage{amsmath,bm}
\usepackage{amsfonts,amssymb}
\usepackage{algorithm}
\usepackage{algpseudocode} 
\usepackage{mathtools}
\usepackage{xspace}
\usepackage{hyperref}
\usepackage{cleveref}
 
\newcommand{\clrb}{\textcolor{black}} 
\DeclarePairedDelimiterX{\infdivx}[2]{(}{)}{%
  #1\;\delimsize\|\;#2%
}

\DeclarePairedDelimiter{\norm}{\lVert}{\rVert}

\algnewcommand{\Or}{\textbf{or}\xspace}

\onlineid{2496}

\vgtccategory{Research}

\vgtcinsertpkg

\title{Efficient Level-Crossing Probability Calculation for Gaussian Process Modeled Data}

\author{Haoyu Li\thanks{e-mail: li.8460@osu.edu}\\ %
        \parbox{1.4in}{\scriptsize \centering The Ohio State University \\ Los Alamos National Laboratory} %
\and Isaac J Michaud\thanks{e-mail: imichaud@lanl.gov}\\ %
     \scriptsize Los Alamos National Laboratory %
\and Ayan Biswas\thanks{e-mail: ayan@lanl.gov}\\ %
     \scriptsize Los Alamos National Laboratory
\and Han-Wei Shen\thanks{e-mail: shen.94@osu.edu}\\ %
     \scriptsize The Ohio State University
     }

\abstract{
Almost all scientific data have uncertainties originating from different sources. Gaussian process regression (GPR) models are a natural way to model data with Gaussian-distributed uncertainties. GPR also has the benefit of reducing I/O bandwidth and storage requirements for large scientific simulations. However, the reconstruction from the GPR models suffers from high computation complexity. To make the situation worse, classic approaches for visualizing the data uncertainties, like probabilistic marching cubes, are also computationally very expensive, especially for data of high resolutions. In this paper, we accelerate the level-crossing probability calculation efficiency on GPR models by subdividing the data spatially into a hierarchical data structure and only reconstructing values adaptively in the regions that have a non-zero probability. For each region, leveraging the known GPR kernel and the saved data observations, we propose a novel approach to efficiently calculate an upper bound for the level-crossing probability inside the region and use this upper bound to make the subdivision and reconstruction decisions. We demonstrate that our value occurrence probability estimation is accurate with a low computation cost by experiments that calculate the level-crossing probability fields on different datasets.%
} % end of abstract

\CCScatlist{
  \CCScatTwelve{Human-centered computing}{Visu\-al\-iza\-tion}{Visu\-al\-iza\-tion techniques}{};
  \CCScatTwelve{Human-centered computing}{Visu\-al\-iza\-tion}{Visualization application domains}{Scientific visualization}
}

\begin{document}

\firstsection{Introduction}

\maketitle

Gaussian processes (GPs) have become an emerging choice \cite{grosskopf2021situ,rumsey2022hierarchical} to represent simulation data in situ to reduce the cost of storage and data I/O. Instead of representing the simulation result as values on structured grids, unstructured grids, or as particles, GPs model the simulation output as a continuous random field defined by a joint distribution of infinitely many random variables. 
Each random variable describes the scalar value distribution at a specific spatial position. Functions and their parameters that define the expected values and the covariances of this joint distribution are called GP priors. Incorporating the GP priors, scalar value distribution at any position can be inferred as a Gaussian distribution conditioned on the saved data samples (observations). 
The fitting of the Gaussian process regression (GPR) model to a scalar field is to find proper priors and the data observations, which is done in situ with the simulation. After the model fitting, the data samples and the GP priors are stored instead of the raw simulation output to reduce the storage and the I/O bandwidth requirement for large simulations.
Compared to other popularly used data reduction approaches, like compression algorithms and neural networks, GPs have the benefit of modeling the uncertainty introduced in the process of simulation and data fitting.

Despite the benefits of data reduction and modeling uncertainties using GPs, visualizing GP-modeled data faces many challenges. 
The first challenge lies in the expensive reconstruction from the GP models.
When we have $m$ data observations, the time complexity to perform dense reconstruction into a grid of $n$ points is $O(m^2n)$. In our experiments, reconstruction from a GPR model with $500$ data observations into a grid of resolution $128 \times 128 \times 128$ takes more than 16 minutes. Scaling to higher resolutions and a larger number of data observations, it can easily take hours or even days to reconstruct from the GPR model.
Even after inferring the GP model in a dense grid, we still face the computation challenge of level-crossing probability (LCP) calculation.
Calculating LCP by probabilistic marching cubes (PMC) is a standard approach to visualize the iso-surface uncertainty. However, PMC is computationally intensive due to the Monte Carlo sampling used to calculate the level-crossing probability. In the original PMC paper, it can easily take ~45 minutes to perform PMC on a relatively small dataset with $256 \times 256 \times 128$ voxels. The computation time scales poorly to higher resolutions because the number of voxels increases cubically with the increase in resolution. 

The computational challenges posed by the Gaussian Process and the probabilistic marching cubes emphasize the need for an efficient method to execute PMC on GP-represented data. 
The computation challenge comes from both the GPR model reconstruction and the Monte Carlo sampling to calculate the probability. Based on the observation that the regions having non-zero level-crossing probability only constitute a small proportion of the total domain, in this paper, we propose an efficient approach to find the non-zero LCP regions, so that the computation is only focused on those regions. 
The basic idea to locate the non-zero LCP region is through a top-down octree subdivision designed and tailored to the GPR data. The whole region is subdivided recursively and only the nodes containing non-zero LCP regions are further subdivided. However, the major difficulty of this approach is how to identify whether a node contains non-zero LCP regions or not. Many previous studies\cite{cignoni1996optimal,livnat1996near, shen1996isosurfacing} on hierarchical iso-surface extraction use the minimum and maximum values of a node to decide whether there is level-crossing inside the node. However, there is no easy way to preprocess a GPR model and get the LCP range for a node. Storing the LCP range information of different nodes also defeats the purpose of using GPR as a succinct data representation.

To tackle the problem, we propose an approach that can efficiently estimate the upper bound of the level-crossing probability for a region. First, we approximate the exact GP with a local GP that only considers the local data observations. This local GP is faster to evaluate and can be used in the regional LCP estimation. 
Second, because most GP priors define a positive covariance matrix, we can find the upper bound of the level-crossing probability of a region by finding the position with the extreme probabilities and assuming the positions inside the region are independent, which avoids the dense reconstruction of data in the region.
Combining these two strategies, we propose a fast algorithm to find the upper bound of the regional LCP and this upper bound can be compared to a threshold to determine whether or not to further subdivide this node. In summary, the contribution of this paper is threefold:
\begin{itemize}
\setlength\itemsep{0em}
    \item We propose a novel and efficient level-crossing probability calculation algorithm from GPR-modeled data through spatial subdivision.
    
    \item We utilize the local GP and statistical rules to efficiently calculate an upper bound for the level-crossing probability, which serves as a foundation for our algorithm.

    \item We demonstrate that the uncertainty visualization from the GPR-modeled data can effectively reveal information about how well the original data are modeled.
    
\end{itemize}

In the rest of this paper, we first survey related studies to the Gaussian processes and uncertain iso-surfaces in \cref{sect:related_works}. Then, we discuss the use of GPR models for data reduction in \cref{sect:gp_background} and the general approach to calculating the LCP on a scalar field represented by a GPR model in \cref{sect:bg-LCP}. We provide the detail of our proposed approach in \cref{sect:method} and evaluate the proposed method using different datasets in \cref{sect:settings} and \cref{sect:results}. Finally, we present an application of our approach in the uncertainty analysis of GPR-represented data in \cref{sect:application}.
We want to emphasize that this paper proposed an approach specifically tackling the problem that the data have already been represented as GP models and we do not have access to the original raw data. In \cref{sect:application}, we compare the GP-modeled data uncertainty and the raw data only to show an application of our approach that increases the efficiency of the uncertainty calculation.

\section{Related Works}
\label{sect:related_works}
Our work focuses on efficient level-crossing probability calculation for GPR models. Therefore, we survey the studies on the use of Gaussian processes in scientific simulations and level-crossing probability fields. 

\subsection{Gaussian Processes in Scientific Simulation and Visualization}
Gaussian processes \cite{williams2006gaussian} are the dominant approach for inference on functions and are used extensively in the analysis of computer experiments \cite{santner2018}. They have recently gained a lot of attention in in-situ data processing to alleviate I/O overhead and reduce the required storage for high-resolution scientific simulations\cite{grosskopf2021situ}. Current studies focus on reducing the time and space complexity of GPs when applied to large data \cite{hensman2013gaussian, liu2020gaussian}. Hierarchical sparse Gaussian processes \cite{rumsey2022hierarchical} are an example that uses Gaussian processes to fit data in-situ generated by large physical simulations. GPs are also utilized for interpolating the data with normally distributed uncertainty and demonstrate superior results compared to linear interpolations\cite{schlegel2012interpolation}. 

Apart from GPs, many studies use other statistical models on scientific data.
Calculating a statistical summary of data blocks is the most straightforward to use statical models on scientific data in situ, which has the advantage of a very small computation overhead. The statistical models used can be either parametric or non-parametric. For example, Gaussian Mixture Models are used by Li et al.\cite{li2020distribution} to reduce particle data and by Dutta et al. \cite{dutta2016situ} and Liu et al. \cite{liu2012gaussian} to model volumetric data. 
In terms of non-parametric models, Thompson et al. \cite{thompson2011analysis} used histograms as a compact representation to represent samples in a local block. All of these mentioned studies separate data into blocks and fit statistical models afterward, which lack the spatial information inside the block and lead to artifacts in reconstruction. 
To solve this problem, Wang et al.\cite{wang2017statistical} propose to save value-based spatial distribution, which combines the spatial and value information inside the block using Bayes' rule.
Rapp et al. \cite{rapp2020visual} model the large multivariate scattered data using Gaussian Mixture models for efficient visualization.

\subsection{Level-crossing Probability Field}

Volume rendering the level-crossing probability field proposed by Pothkow and Hege \cite{pothkow2010positional} is the ubiquitous approach to visualize the iso-surfaces with uncertain data. They set up the theoretical foundations for the uncertain iso-surfaces in the paper and derive the equations for the level-crossing probability.
In a follow-up work\cite{pothkow2011probabilistic}, they propose the probabilistic marching cubes as an algorithm to generate LCP fields for the Gaussian random field with correlation considered. 
However, they rely on Monte Carlo sampling to calculate the LCP in the paper, which makes the algorithm very computationally expensive. To solve the computation issue, Pothkow et al.\cite{pothkow2013approximate} approximate the cell level-crossing probability with the maximum of the edge-crossing probability. Pfaffelmoser et al. \cite{pfaffelmoser:2011:VPGV} propose an incremental way to calculate the first level-crossing probability in a correlated random field through volume ray-marching, which reduces the computation cost compared to Monte Carlo sampling. \clrb{However, their method requires a special condition on the correlation which most GP random fields cannot satisfy.}

The LCP calculation is also generalized to the uncertainty of other types of distributions. Athawale et al. derived the closed-form equations for LCP calculation on the random field with independent uniformly distributed uncertainties\cite{athawale2013uncertainty} and nonparametric distributions\cite{athawale2015isosurface, athawale2020direct}. 
Gillmann et al. also use the multi-variate uncertainty model to analyze the uncertainties from geometric data \cite{gillmann2018modeling} and the surfaces in the image data \cite{gillmann2018accurate}.
Despite the efficient closed-form calculations for different types of distributions in these papers, it is still assumed that uncertainties in the random field have no spatial correlation. Calculating the exact cell-crossing probability for a correlated random field is still a challenging problem. 
A recent study solves the computation challenges using neural networks\cite{han2022accelerated}. However, the accuracy of the predicted probability largely depends on the training dataset, and the generalizability of the prediction model is unknown. 

In this paper, we propose a novel spatial subdivision approach to calculate the level-crossing probability on a random field with a correlation defined by a Gaussian process. The efficiency is gained by only calculating the exact LCP in the cells with non-zero probability. We use a fast estimate of the LCP in the large region to help identify the regions with non-zero probability.

\section{Gaussian Process Regression for Data Reduction}
\label{sect:gp_background}
In this section, we first introduce the sparse Gaussian process regression (SGPR) model and describe how it is applied to the data reduction problem. 
The scalar field from a scientific simulation is defined as a functional mapping from a spatial location $\bm{x}$ to a scalar value $y$. 
The Gaussian process models this scalar field as a continuous random field, where the value at each different spatial location is a random variable $Y$. 
The random variables in the random field are modeled as a multidimensional joint Gaussian distribution. 
The expected value of each variable is defined by a prior mean function, while the correlation between different random variables is defined by a kernel function applied on the spatial location $\bm{x}_1$ and $\bm{x}_2$ of any two random variables: $k(\bm{x}_1,\bm{x}_2)$. 
For example, a popularly used radial basis kernel function (RBF) defines the covariance between $\bm{x}_1$ and $\bm{x}_2$ as $\sigma^2e^{-\norm{\bm{x}_1-\bm{x}_2}^2/\left(2\ell^2\right)}$, where $\sigma$ and $\ell$ are parameters for this kernel function.
Since the multivariate Gaussian distribution is defined between arbitrary spatial locations (prior), given the scalar value at some places (observations), we can calculate the conditional distribution of scalar values at other locations (posterior). The process of inferring unknown points given the observations is called the Gaussian process regression (GPR).
Applying the Gaussian process regression model, domain scientists can reduce the scalar field data represented by dense grids to a sparse subset of positions and the parameters of the multivariate Gaussian prior distribution, which describes the relations between the positions.

We follow the previous studies \cite{grosskopf2021situ,rumsey2022hierarchical} and use the technique called the sparse Gaussian process regression (SGPR) to choose data samples. SGPR introduces the concept of the inducing points as data observations. Inducing points can be placed in any spatial location in the domain so that the posterior calculated using the inducing points is similar to the posterior calculated using the full dataset. Using the inducing points reduces the storage cost of the data and the computation time to infer the posterior. The SGPR model finds the inducing points' positions using a likelihood maximization approach. Inducing points' positions are chosen to maximize the probability of the existent data observations, which, in our case, are the values on the grid. We will not dive into the details of SGPR model fitting and suggest the original paper \cite{snelson2005sparse} that explains this in detail. Briefly, apart from the parameters of the inducing points' positions, the SGPR model's fitting process also finds the parameters for the covariance matrix kernel function $k$ and a parameter $\sigma_y$ that describes the measurement error. If we are training a 3D model SGPR model with a kernel that has two parameters (e.g. RBF kernel),  in total there is $3m + 3$ number of parameters to store, where $m$ is the number of inducing points. The original dataset with $n$ grid points requires the storage of $3n$ numbers. Replacing the original data with the SGPR model will significantly reduce the storage cost since the number of inducing points $m$ is significantly less than the number of grid points $n$. Next, we explain how the posterior is inferred given all the parameters in an SGPR model.

The GP prior, i.e., the mean value function and the kernel function, essentially describe the joint Gaussian distribution among the random variable at an arbitrary set of locations. Therefore, once we know the inducing points' positions, given the training data at the grid points, we can follow the conditioning rules for the multidimensional Gaussian and the Bayes rules to infer the joint Gaussian distribution of the random variables at the inducing point positions. 
The parameters for this joint distribution are saved after the model training and used later when reconstructing from the SGPR model.
We denote the set of all inducing points as $M$ and this distribution as $\mathcal{N}(\mu_M, A_M)$, where $\mu_M \in \mathbb{R}^{m}$ is the mean vector and $A_M \in \mathbb{R}^{m\times m}$ is the covariance matrix for the inducing point distribution.
The detailed derivation for $\mu_M$ and $A_M$ can be found in \cite{snelson2005sparse}.

Finally, given the inducing points' distribution, we can infer the posterior following the conditioning rules of the multivariate Gaussian distributions. For a set $I$ containing the positions of interest for inference at locations $x_i \in I$. The values at these positions follow a Gaussian distribution conditioned on the inducing point set $M$: $\mathcal{N}(\mu_{I|M}, \Sigma_{I|M})$. 
We use a short-hand notation $K_{IJ}$ to denote the kernel function evaluated between two sets of positions. For example, the i-th row and j-th column of this matrix $[K_{IJ}]_{ij}$ is calculated using the kernel function as $k(\bm{x}_i,\bm{x}_j)$, where $i$ and $j$ iterate through the elements in the set $I$ and the set $J$. Following this notation, we can infer the posterior distribution of the points in the set $I$ given the inducing points in $M$ as:
\begin{equation}
\begin{aligned} 
\label{eq:sgp}
    \mu_{I|M}    & = K_{IM}K_{MM}^{-1} \mu_M \\
    \Sigma_{I|M} & = K_{II} - K_{IM}K_{MM}^{-1}K_{MI} + K_{IM}K_{MM}^{-1}A_{M}K_{MM}^{-1}K_{MI},
\end{aligned}
\end{equation} 
where $\mu_M$ and $A_M$ are already calculated and stored in the SGP fitting.
In this equation, only matrices $K_{IM}$, $K_{MI}$, and $K_{II}$ are variables determined by the inference positions $\bm{x}_i \in I$, other values are known and can be precomputed in inference stage. 
In summary, $\mu_{I|M}$ and $\Sigma_{I|M}$ of this distribution can be written as a function of $\bm{x}_i \in I$. 

This equation determines the computation complexity of the SGPR model inference. With proper precomputation, the time complexity to calculate $\mu_{I|M}$ is $O(m)$, and the time complexity to calculate $\Sigma_{I|M}$ is $O(m^2)$. Therefore, when we are performing a dense reconstruction from the SGPR model back to the original grid data with $n$ grid points. The overall time complexity is $O(m^2n)$.
\clrb{High complexity prevents us from performing dense reconstruction from the SGPR model, which motivates us to reduce the number of needed point reconstructions $n$ through hierarchical subdivision and the time for each reconstruction through local SGP approximation.}

\section{Level-crossing Probability for GP Random Field}
\label{sect:bg-LCP}

In this section, we discuss how the conditional distribution inferred from the SGPR model is used to calculate the level-crossing probability (LCP). According to the paper by Pöthkow et al \cite{pothkow2010positional}, the level-crossing probability for a line segment $[\bm{x}_j, \bm{x}_k]$ can be calculated as:
\begin{equation}
\begin{aligned}
    P(\theta\text{-crossed}) = 1& - P(Y_j<\theta \cap Y_k<\theta)  \\
                                 & - P(Y_j>\theta \cap Y_k>\theta),
\end{aligned}
\end{equation}
where $Y_j$ and $Y_k$ are random variables defined at $\bm{x}_j$ and $\bm{x}_k$.
This calculation assumes that the realizations between the sample points are linearly interpolated, which is natural for the dataset discretized into a grid. However, more discussion is needed for a continuous random field described by a GP model. 

Theoretically, a line segment in the GP random field is modeled as a joint Gaussian distribution of an infinite number of random variables at positions in this line segment. The calculation of the level-crossing probability of this line segment requires us to calculate $1 - P(Y_0<\theta \cap ... \cap Y_i<\theta \cap ...) - P(Y_0>\theta \cap... \cap Y_i>\theta \cap ...)$, where $i=1, ..., \infty$ denotes all the random variables inside the line segment. Unfortunately, this probability cannot be calculated directly. However, for a small enough line segment, realizations from the GP model can be assumed to be linear, which means we can use the random variables at the line segment edges to calculate the level-crossing probability. In the 3D case, to calculate the LCP field from an SGPR model, we need to discretize the field into a grid. The grid resolution should be at least the resolution of the original grid data that we use to fit the SGPR model. Then, we infer the random variables' joint distribution at the corners of the cells in the grid and use that to finally calculate the LCP for the cell. 
In this discretization, although the target resolution for the finest cell is fixed, this does not enforce that we need to use the same sample distance at any location in the domain. For example, it is very common for a large region far from the iso-surface to have zero LCP in practice. This observation motivates us to use an adaptive sample approach for the LCP field calculation from the SGPR model.

Next, we introduce a straightforward baseline approach to calculate the LCP field from an SGPR model, where we discretize the field regularly and densely into a grid and calculate the LCP for each small cell. For each small cell, the values at eight corners of the cell follow an 8-variate Gaussian distribution, whose mean and covariance can be calculated using \cref{eq:sgp}. The $\theta$-level-crossing probability for this cell is then defined as:
\begin{equation*}
    P(\theta\text{-crossed}) = 1 - P(\theta\text{-not-crossed}).
\end{equation*}
The probability $P(\theta\text{-not-crossed})$ is the sum of the probability that all eight corner values are either lower or higher than the iso-value:
\begin{equation*}
\label{eq:theta_non_crossing}
\begin{aligned}
    P(\theta\text{-crossed}) = & P (Y_1<\theta,Y_2<\theta,...,Y_8<\theta) + \\
                                    & P (Y_1>\theta,Y_2>\theta,...,Y_8>\theta) \\
                                  = & \idotsint\limits_{y_1<\theta \wedge ... \wedge y_8<\theta} f(y_1,...,y_8) \,dy_1...\,dy_8 + \\
                                    & \idotsint\limits_{y_1>\theta \wedge ... \wedge y_8>\theta} f(y_1,...,y_8) \,dy_1...\,dy_8 ,
\end{aligned}
\end{equation*}
where $f(y_1,...,y_8)$ denotes the probability density function for the 8-variate Gaussian distribution. However, this multi-variable integration cannot be calculated in the closed form. Monte Carlo (MC) sampling can be used to estimate this probability, i.e. we can sample this joint Gaussian distribution and use the percentage of the samples where the level-crossing occurs to approximate the level-crossing probability. We need to determine the number of samples when performing the MC sampling. More samples will give more accurate results however require more computation.

In this dense reconstruction baseline approach, because both GP inference of the cell corner distributions and the Monte Carlo sampling are expensive, the total computation time is long even for a relatively low resolution. In the next section, we propose our approach to tackle this problem. 

\section{Method}
\label{sect:method}

From \cref{sect:bg-LCP}, we can see that it is possible to densely discretize a random field represented by an SGPR model into small cells and then calculate the probability for each cell through Monte Carlo sampling. As long as the discretization is dense enough, the resulting level-crossing field will approximate the true field well. However, as discussed before, the issue with dense discretization is that the computation for SGPR reconstruction and Monte Carlo sampling are both very expensive. Therefore, we propose an adaptive spatial subdivision approach to efficiently calculate the LCP field, which reduces the cost of computation by only densely discretizing the regions with non-zero probability.

The core idea of our method is to skip regions that have zero level-crossing probability efficiently before dense reconstruction in the region, which is achieved by calculating the upper bound of the \textit{regional} level-crossing probability, explained in detail in this section. Using this upper bound, we subdivide the space adaptively following an octree structure. If the regional level-crossing probability of a node is less than a set threshold, we do not need to subdivide the node anymore. Otherwise, the nodes are subdivided until we hit a predetermined level, which corresponds to the smallest sample distance of our choice. The algorithm of the subdivision is also presented in this section.

\subsection{Estimation of the Regional Level-Crossing Probablity}
\label{sect:prob_estimation}

We construct the regional level-crossing probability upper bound by computing a lower bound on $P(\theta\text{-not-crossed}) = 1- P(\theta\text{-crossed})$. 
Assuming we already determine a proper discretization density for the region and this discretization results in a total of $d$ grid points inside the region, the $\theta$-not-crossed probability for this hypothetical grid is calculated as $P(\theta\text{-not-crossed}) = P(Y_1<\theta,..., Y_d<\theta) + P(Y_1>\theta,..., Y_d>\theta)$. We will compute a lower bound for each of these probabilities separately to determine the upper bound for $P(\theta\text{-not-crossed})$. 
Using the assumption that these $d$ grid points follow a d-variate Gaussian distribution with an isotropic correlation function, the probability that all values in the block are less than $\theta$, $P(Y_1<\theta,..., Y_d<\theta)$, is easily bounded using Slepian's inequality \cite{hsu1996multiple}
\begin{equation}\label{slepian} 
    P(Y_1<\theta,\ldots,Y_d<\theta) \geq \prod_{i=1}^d P(Y_i < \theta).
\end{equation}
The right-hand side of (\ref{slepian}) is further bounded below by $B_\ell^d$ where 
\begin{equation*}
    B_\ell^d = [\min_{i \in D} P(Y_i < \theta)]^d.
\end{equation*}
The set $D$ contains all grid points inside the region of interest.
The lower bound for $P (Y_1>\theta,Y_2>\theta,...,Y_d>\theta)$ is calculated similarly, with $B_u^d$, where
\begin{equation*}
   B_u^d = [\min_{i \in D} P(Y_i > \theta)]^d.
\end{equation*}
Therefore the upper bound on the iso-surface crossing probability is
\begin{equation}
    \label{eq:upper_bound}
    P(\theta\text{-crossed}) \leq 1 - B_\ell^d - B_u^d.
\end{equation}
This equation suggests that to calculate this upper bound, instead of considering all $d$ grid points, we can simply find the minimum probability inside the region of interest. Furthermore, this random variable with the minimum probability does not need to come from $d$ discrete samples in the region. Because the discrete samples are only a subset of all the points in the region, we can instead find any random variable $Y$ continuously defined in the region that minimizes the probability $P(Y<\theta)$ and $P(Y>\theta)$. 
Next in this section, we discuss how to find this minimum probability.

The key idea of our method is based on the fact that any random variable $Y$ at any position $\bm{x}$ inside the region follows a Gaussian distribution $\mathcal{N}(\mu,\sigma^2)$. Therefore, the probability $P(Y < \theta)$ is calculated as  $\Phi \left (\frac{\theta - \mu} {\sigma}\right )$, where $\Phi$ is the cumulative density function for a standard Gaussian distribution.
We also know that following \cref{eq:sgp}, when there is only one element $\bm{x}$ in the set $I$, $\mu_{\{\bm{x}\}|M}$ and $\Sigma_{\{\bm{x}\}|M}$ are solely dependent on the position $\bm{x}$, so that both $\mu = \mu_{\{\bm{x}\}|M}$ and $\sigma^2 = \Sigma_{\{\bm{x}\}|M}$ can be written as a function of the location $\bm{x}$. 
For convenience, we denote this target function as:
\begin{equation*}
    P(Y<\theta) = F_M(\bm{x}) \coloneqq \Phi \left (\frac{\theta - \mu_{\{\bm{x}\}|M}} {\Sigma_{\{\bm{x}\}|M}}\right ),
\end{equation*}
which returns the probability value for the random variable $Y$ at location $\bm{x}$ given the inducing point set $M$ from the SGPR model. Similarly, $P(Y>\theta) = 1 - F_M(\bm{x})$. Then, the problem of finding $B_\ell$ and $B_u$ is converted to finding the minimum values for the functions $F_M(\bm{x})$ and $1 - F_M(\bm{x})$. 

Since the gradient of this function can also be calculated in closed form, we find the minimum values using a quasi-Newton method L-BFGS-B\cite{liu1989limited}. After finding $B_\ell$ and $B_u$ we can easily calculate the upper bound for $P(\theta\text{-crossed})$ following \cref{eq:upper_bound}.

\subsection{Local SGP Approximation}
\label{sect:local_sgp}
%local SGP approximation
Compared to densely reconstructing $d$ grid points inside the region, we reduce the number of SGPR predictions from $d$ to the number of iterations in the L-BFGS-B algorithm. 
As mentioned in \cref{sect:gp_background}, the time complexity for the inference on a single position is quadratic to the number of inducing points $m$. The complexity is still high when the number of inducing points is large. Therefore, to further reduce the computation time, we propose to only use local inducing points to approximate the full SGPR model. 

The construction of the local SGPR model is to remove the inducing points with too small covariance at the position of prediction.
Kernel functions used by SGPR models define the covariance concerning the distance between the two points. For example, the most widely used kernel RBF defines the covariance as 
\begin{equation*}
    \text{RBF}(\bm{x}_1,\bm{x}_2)=\sigma^2e^{-\norm{\bm{x}_1-\bm{x}_2}^2/\left(2\ell^2\right)}
\end{equation*}
where $\sigma$ and $\ell$ are hyperparameters for this kernel. The covariance decreases when the distance increases. With the hyperparameter known for the kernel, setting a threshold for the covariance, we can easily calculate a threshold $\beta$ for the distance, so that points having distances greater than $\beta$ will have covariance less than the set threshold. This distance threshold can always be found when the kernel function is isotropic and monotonic, which covers most of the Gaussian process kernel functions. 

\begin{figure}[htb]% specify a combination of t, b, p, or h for the top, bottom, on its own page, or here
  \centering % avoid the use of \begin{center}...\end{center} and use \centering instead (more compact)
  \includegraphics[width=\columnwidth]{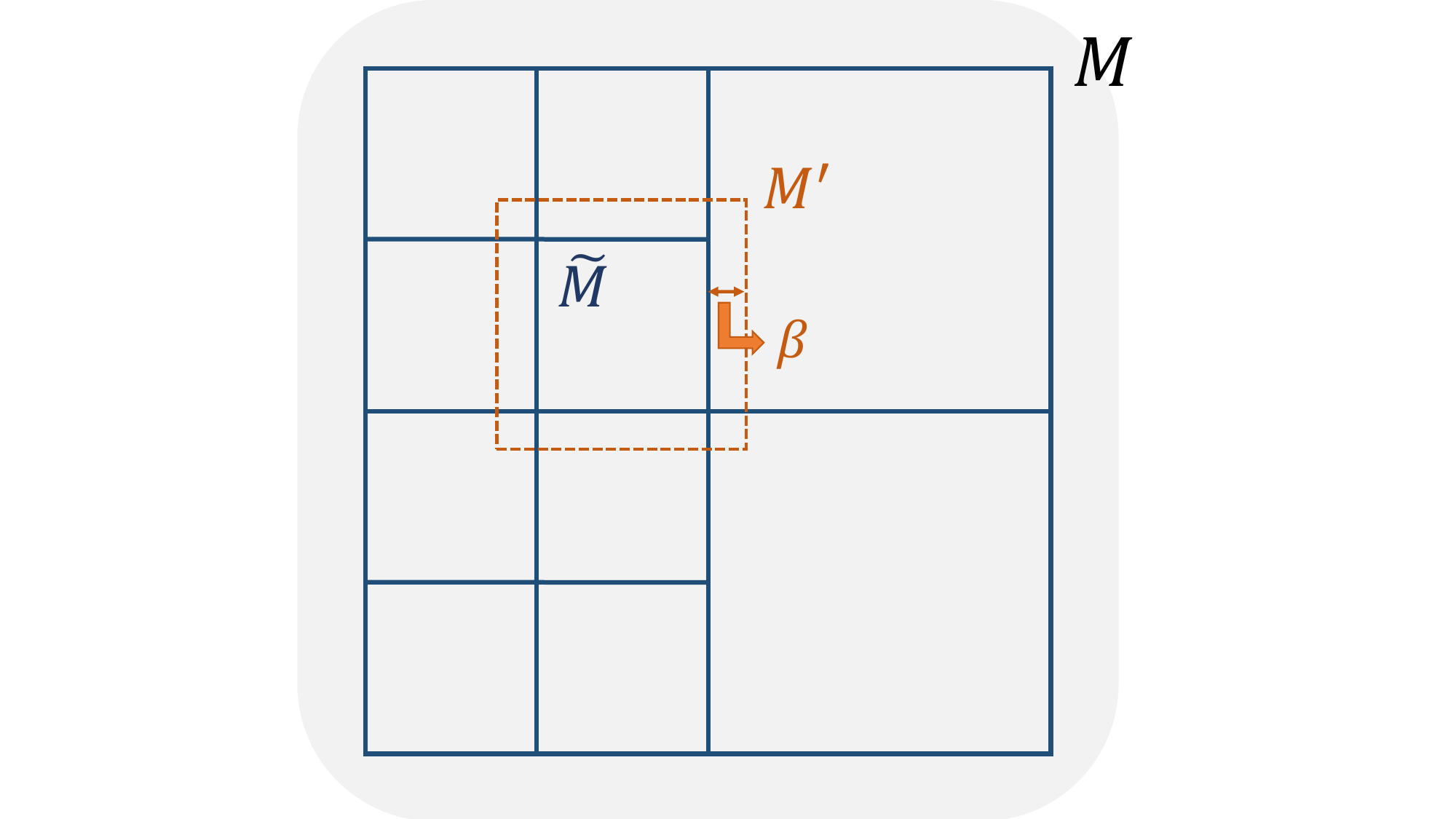}
  \caption{%
    We show a 2D diagram of the regions defining different subsets of the inducing points. $M$ is the set containing all the inducing points in the domain. $\Tilde{M}$ is the set of inducing points in the region of a specific node. $M'$ is the set of inducing points in the enlarged node region (orange dashed square) by the threshold $\beta$. 
  } 
  \label{fig:octree}
\end{figure}

To calculate the upper bound for the regional level-crossing probability, we enlarge this region by the distance threshold $\beta$ and use the local inducing point set $M'$ inside this enlarged region to build the local SGPR model. 
A diagram of this enlarged region is shown in \cref{fig:octree}.
Then, instead of finding the minimum probability for $P(Y<\theta) = F_M(\bm{x})$, we find the minimum probability for $F_{M'}(\bm{x}) \approx F_M(\bm{x})$, which greatly reduce the computation for the L-BFGS-B algorithm.
It is worth noting that this local approximation is only used when we are estimating the regional level-crossing probability. To calculate the level-crossing probability for a small cell, we still use the original full SGPR model to infer the multivariant Gaussian distribution among their corner points. 

\subsection{Octree Subdivision}
\label{sect:octree_subdivision}
An octree is built top-down to subdivide the space for our algorithm.
The whole spatial domain is first represented by a single root node. 
We recursively subdivide the node by comparing a probability threshold $\alpha$ (usually slightly larger than 0) and the estimated regional level-crossing probability (explained in \cref{sect:prob_estimation}) of the node until no node is subdivided or we reach the maximum level that is predefined. 
If the upper bound calculated for the block is smaller than $\alpha$, the iso-surface crossing probability will also be less than $\alpha$ and we can safely skip the block. Otherwise, the block is subdivided. 
We first present our procedure in \cref{alg:subdivision} and then discuss the details of the procedure.

\begin{algorithm}[ht]
	\caption{Octree Adaptive Query} 
    \label{alg:subdivision}
	\begin{algorithmic}[1]
    \Require {A GP model with inducing points $M$, domain bounds $\bm{x}_\ell,\,\bm{x}_u \in \mathbb{R}^3$, iso-value $\theta$, and thresholds $\alpha$  and $\beta$. Initially, $M' = M$.}
    \Ensure {Predicted set $S$ of the non-zero probability nodes.}
\Procedure{Octree}{$M', \bm{x}_l, \bm{x}_u$}
    \State $\text{subdivide} \gets \textbf{False}$
    \State $ \Tilde{M} \gets \{\bm{x}_i \,|\, \bm{x}_i \in Box(\bm{x}_l, \bm{x}_u)$ and $ \bm{x}_i \in M\}$
    \State $p_1 \gets max_{\bm{x}_i \in \Tilde{M}}{P(Y_i<\theta)}$
    \State $p_2 \gets max_{\bm{x}_i \in \Tilde{M}}{P(Y_i>\theta)}$
    \If{$ p_1>\alpha \, \textbf{and} \, p_2>\alpha$}
        \State $\text{subdivide} \gets \textbf{True}$ 
    \Else
        % \State $\bm{x}_c \gets (\bm{x}_l + \bm{x}_u) /2$
        % \Comment{Node Center}
        % \State $ M' \gets \{\bm{x}_i \,|\, \norm{\bm{x}_c-\bm{x}_i} < \beta,\, \bm{x}_i \in M\}$
        
        \If {$p_2<\alpha$}
            \State $B \gets \min{F_{M'}(\bm{x})}$
            \Comment{Using Local GP (\cref{sect:local_sgp})}
        \EndIf
        \If {$p_1<\alpha$}
            \State $B \gets \min{\left ( 1 - F_{M'}(\bm{x})\right ) }$
        \EndIf
        \State $d \gets$ \# Grid points for this level
        \If {$1-B^d > \alpha$} 
        \Comment{Compare to the upper bound}
            \State $\text{subdivide} \gets \textbf{True}$ 
        \EndIf
    \EndIf
    \If {subdivide}
        \If {Max depth reached}
            \State \Return $\{Box(\bm{x}_l, \bm{x}_u)\}$
        \Else
            \State $S_{\text{ret}} \gets \emptyset$
            \For {$i \gets 1$ to $8$}
                \State $\bm{x}'_u, \bm{x}'_l \gets\text{Child}(\bm{x}_u, \bm{x}_l, i)$
                \Comment{Bounds for i-th child}
                \State $M'_{Child} \gets \{\bm{x}_i | \, \bm{x}_i \in M', \bm{x}_i \in Box(\bm{x}'_u -\beta, \bm{x}'_l + \beta)\}$
                % \Comment{Inducing points for child}
                \State $S_{\text{ret}} \gets S_{\text{ret}} \cup \text{Octree}(M'_{Child}, \bm{x}'_l, \bm{x}'_u)$
            \EndFor
            \State \Return $S_{\text{ret}} $
        \EndIf
    \Else
        \State \Return $\emptyset$
    \EndIf
 \EndProcedure
	\end{algorithmic} 
\end{algorithm}

Since it still requires some computation to numerically find the extrema using L-BFGS-B, we use the inducing points inside the node to make a preliminary decision first and avoid finding the extreme values in some situations.
As mentioned in \cref{sect:gp_background}, the inducing points' values follow a Gaussian distribution. The parameters for the distribution are known after the model fitting. In the process of adaptive query, we can arrange the inducing points in different nodes and treat them as known data samples. 
If there already exist some data samples that indicate a non-zero level crossing probability in the node region, we can skip some upper bound calculation steps or directly subdivide the node.
We represent the inducing points inside a specific node as set $\Tilde{M}$ as shown in \cref{fig:octree}, which is a subset of all the inducing points $M$.
Given the iso-value $\theta$, we can calculate the probability that inducing points' value is larger than or smaller than $\theta$ using the mean vector $\mu_{\Tilde{M}}$ and their covariance matrix $A_{\Tilde{M}}$. We denote the maximum of these two probabilities as $max_{\bm{x}_i \in \Tilde{M}}{P(Y_i<\theta)}$ and $max_{\bm{x}_i \in \Tilde{M}}{P(Y_i>\theta)}$.
Depending on the values of these two probabilities, we make different decisions in the following three different cases: 
\begin{itemize}
    \item $max_{\bm{x}_i \in \Tilde{M}}{P(Y_i<\theta)}$ and $max_{\bm{x}_i \in \Tilde{M}}{P(Y_i>\theta)}$ are larger than $\alpha$. 
    \item $max_{\bm{x}_i \in \Tilde{M}}{P(Y_i>\theta)}$ is smaller than $\alpha$.
    \item $max_{\bm{x}_i \in \Tilde{M}}{P(Y_i<\theta)}$ is smaller than $\alpha$.
\end{itemize}
It is worth noting that because ${P(Y_i<\theta)} = 1 - {P(Y_i>\theta)}$, and $\alpha$ is chosen to be slightly larger than 0, it is not possible both values are smaller than $\alpha$.

In the first situation, because there are non-zero probabilities for inducing points' values larger than and smaller than the iso-value, according to the definition of the level-crossing, it is very likely the level-crossing probability is non-zero. Therefore, we can directly subdivide the node without upper bound estimation. 
In the second case, since the maximum probability that the inducing points are larger than the iso-value is very small, $B_u$ in \cref{eq:upper_bound} will be even smaller and is negligible in calculating the upper bound. In the third case, similarly, $B_\ell$ is negligible. 
We save the computation for finding both extrema in the first situation and the computation to find one extrema in the second and the third situation.

From lines 9 to 18 of \cref{alg:subdivision}, first, we use the method discussed in \cref{sect:local_sgp} to select local inducing points to further reduce the computation cost. It is worth noting that the local inducing points $M'$ is a superset of the inducing points inside the node $\Tilde{M}$ as shown in \cref{fig:octree}. 
We store a copy of the corresponding $M'$ in each node in the process of subdivision to make the query of $M'$ and $\Tilde{M}$ fast.
Then, we calculate the upper bound under cases two and three using the methods discussed in \cref{sect:prob_estimation} and compare the upper bound to the threshold $\alpha$. 

Finally, the procedure will output a collection of predicted nodes that have non-zero level-crossing probability. We only need to reconstruct the Gaussian distribution among the corners for these nodes from the SGPR model and then use Monte Carlo sampling to get the LCP. Other region's LCP values are assigned to zero.

%-------------------------------------------------------------------------
\section{Experiment Settings}
\label{sect:settings}
In this section, we explain how we conducted our experiments to evaluate our algorithm. 
Because the approach proposed in this paper is to accelerate the level-crossing probability calculation on GPR-modeled data, the most important things to evaluate are the calculation time and the accuracy. In addition, we also evaluate the scalability, hyperparameter choice, and adaptive query pattern of our approach. 
Apart from evaluating our method, we also perform the uncertainty analysis on the experiment datasets to demonstrate the application of our approach.
Below we first give the details about our implementation, the metrics, the baselines, and the dataset in the experiments.

\subsection{Implementation}
We implemented the algorithm mentioned above using C++, Cython, and Python. The GPy\cite{gpy2014} framework is used to train the sparse Gaussian process regression model. We use Cython\cite{behnel2011cython} to integrate our implementation in C++ and the GPR model trained using a Python framework. Our source code are available at \href{https://osf.io/wpgyb/?view_only=c9d6b1cdae694f83994244c6edc5eca7}{osf.io (link)}.

\subsection{Metrics}
As mentioned before, we want to evaluate the calculation time and the accuracy of our approach. The calculation time is simply measured in seconds. We separate the time for GPR inference, Monte Carlo sampling, and the overhead of adaptive query in the comparison. In terms of LCP accuracy, since our approach calculates an approximated LCP field by assuming the probability that is less than a threshold to be zero, there could be a difference between our extracted LCP and the LCP extracted through dense grid reconstruction and Monte Carlo sampling (ground truth). We quantify this error using the root mean squared error (RMSE) between the LCP field from our approach and the ground truth to show the accuracy of our approximation. We also compare the LCP field through volume rendering to evaluate the methods qualitatively.
To evaluate the adaptive query pattern of the approach, we volume render the octree level of our query. Ideally, only the regions with non-zero probability should have a deep level of subdivision. 

\subsection{Datasets}
We used two datasets to evaluate our approach. The first one is a synthesized dataset called Tangle. The function defining this dataset can be found in the paper by Knoll et al.\cite{knoll2009fast}. We are interested in the surface of the iso-value $-0.59$. This surface has thin bridges across different large components, which is useful to show the effect of uncertainty. 
For Tangle, the dataset is first discretized into a grid of size $32 \times 32 \times 32$ and we fit a sparse Gaussian process regression (SGPR) model with $50$ inducing points.

The second dataset we used is the ethanediol dataset from the topology toolkit \cite{tierny2017topology}. Different iso-values can reveal various interesting structures of an ethanediol molecular from this dataset. We experiment on different iso-values that correspond to iso-surfaces of various shapes and densities, but we are most interested in the iso-value $-1.5$ in the uncertainty analysis. The ethanediol dataset comes in a grid of $115 \times 116 \times 134$ and we train an SGPR model with $500$ inducing points. Both models for the two datasets use the RBF kernel, and the detailed SGP hyperparameters are presented in \cref{tab:gp_para}. It is worth noting that these SGP hyperparameters are obtained through the fitting process (different from the hyperparameters $\alpha$ and $\beta$ which are hand-picked by us). 

\begin{table}[ht]
  \caption{
  The hyperparameters of the SGPR models we test. PSNR is the metric to quantify the SGPR model reconstruction quality and the storage column shows the relative storage cost of the SGPR model compared to the grid data for training.
  }
  \label{tab:gp_para}
\begin{tabular*}{\columnwidth}{@{\extracolsep{\fill}}c c c c c c}
  \hline
     &$\ell$  & $\sigma$  & $\sigma_y$ & PSNR & Storage  \\
  \hline 
 Tangle  & 36.33 & 187.6 & 0.0015 & 32.90 dB & 4.43\% \\
 Ethanediol  & 12.79 & 0.0036 & 0.0061 & 41.28 dB & 7.10\% \\
  \hline 
\end{tabular*}
\end{table}

\subsection{Baseline}
The baseline we compare in the evaluation is densely reconstructing the SGRP model into a grid and then calculating the LCP field through Monte Carlo (MC) sampling. As mentioned in the introduction, our approach uses an adaptive query scheme to reconstruct and perform MC sampling at places of non-zero probability. We reduce the total number of cells that perform reconstructions and MC sampling. Approaches\cite{pothkow2013approximate} that reduce the cost of MC sampling could be combined with our approach to further reduce the calculation. Our method is still valid to reduce the cost of GP reconstruction in that case.

Another setting worth mentioning for the evaluation is the number of samples in the Monte Carlo sampling. We choose $1600$ which is relatively small compared to the ones used by previous studies\cite{han2022accelerated, pothkow2011probabilistic}. If the number of samples increases, the time reduction of our method should be larger than the results reported in the paper.

%-------------------------------------------------------------------------
\section{Results}
\label{sect:results}

In this section, we report our evaluation results in terms of computational performance, accuracy, scalability, hyperparameter choice, and adaptive query patterns. 

\subsection{Iso-surface Extraction Time}
\label{sect:time}
\begin{figure}[htb]% specify a combination of t, b, p, or h for the top, bottom, on its own page, or here
  \centering % avoid the use of \begin{center}...\end{center} and use \centering instead (more compact)
  \includegraphics[width=\columnwidth]{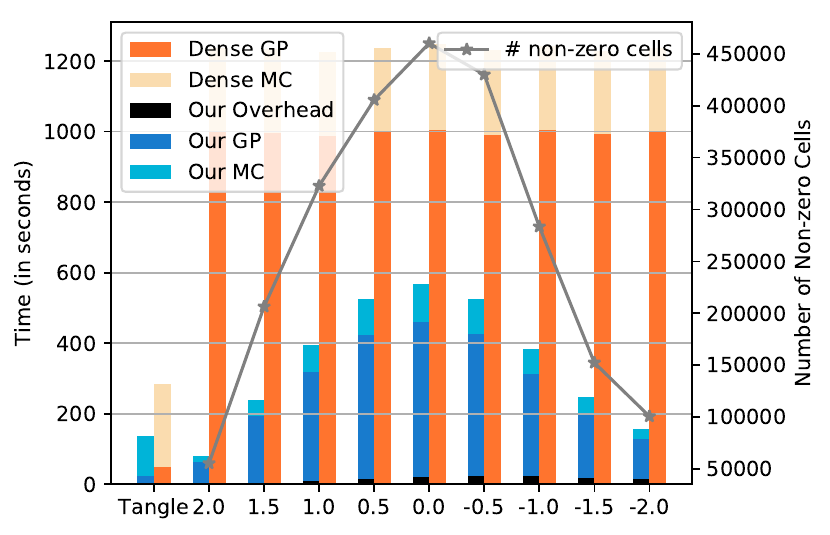}
  \caption{%
    LCP computation time comparison between our method and the baseline method for two datasets. The first column shows the results for the tangle dataset and the other columns show the results for different iso-values in the ethanediol dataset. For the ethanediol dataset, we use a line chart showing the number of non-zero cells in the ground truth LCP field.
    The computation time is separated into GP inference time, Monte Carlo sampling time, and the overhead to perform the adaptive query in our approach. As can be seen, the computation time for our approach is proportional to the number of non-zero probability cells but the baseline method is not sensitive to the number of iso-surface cells and hence is more expensive. 
  } 
  \label{fig:time}
\end{figure}

% time
For the runtime evaluation for our approach, we choose the target resolution of $128\times128\times128$ for both datasets.
The calculation time of our approach and the baseline are shown in \cref{fig:time}. The time for the overhead, the GP reconstruction, and the MC sampling are separated in the figure using a stacked bar of different colors. 
For the tangle dataset, we test the iso-value $-1.5$.
For the ethanediol dataset, we test iso-values in the range from $-2.0$ to $2.0$ to cover the LCP field with different numbers of non-zero probability cells.

First, we can see a clear reduction in the computation time of our approach compared to the baseline. 
The total computation time of our approach is $48.6\%$ of the dense reconstruction for the tangle dataset and ranges from $6.4\%$ (iso-value equals 2.0) to $45.8\%$ (iso-value equals 0.0) for the ethanediol dataset. 
The time for our approach depends on the number of cells estimated to have non-zero probability, which depends on two factors. The first is how accurate our probability estimate is and the second is the actual number of non-zero probability cells in the volume. 
When we have accurate regional probability estimates (tight upper bounds), we can reduce the number of subdivisions in the adaptive query and the time for the GP reconstruction and the Monte Carlo (MC) sampling.
Similarly, if the actual number of non-zero probability cells is smaller, the adaptive query overhead, the time for GP, and the time for MC are all reduced.
To confirm the number of non-zero probability cells indeed influences our approach's computation time, in \cref{fig:time}, we plot the actual number of non-zero probability cells in comparison to our computation time. It can be observed that when the actual number of non-zero probability cells increases when the iso-value is around 0.0, the computation time for our approach also increases. For the accuracy of our regional probability upper bound estimation, we visualize the adaptive query pattern of our technique in \cref{sect:query_pattern}.

The baseline method runtime is different between the two datasets, because of the different number of inducing points for these two SGP models, and thus dense GP reconstruction time differs a lot. For the tangle dataset (50 inducing points), dense GP reconstruction into a grid of $128 ^ 3$ takes about $48$ seconds, while for the ethanediol dataset, it takes about $1000$ seconds. We will further discuss the impact of the model size in \cref{sect:scalability}.
On the other hand, the runtimes for the Monte Carlo sampling in the baseline are similar across datasets because we use the same amount of samples ($1600$). A larger number of Monte Carlo samples can be used, in which case we can expect more runtime reduction using our approach.

In summary, our approach reduces the computation time through an adaptive query, the amount of which depends on the actual number of non-zero probability cells. 
Next, we are going to evaluate the accuracy of the LCP field of our approach.

\subsection{Accuracy}
\label{sect:acc}
\begin{figure}[htb]% specify a combination of t, b, p, or h for the top, bottom, on its own page, or here
  \centering % avoid the use of \begin{center}...\end{center} and use \centering instead (more compact)
  \includegraphics[width=\columnwidth]{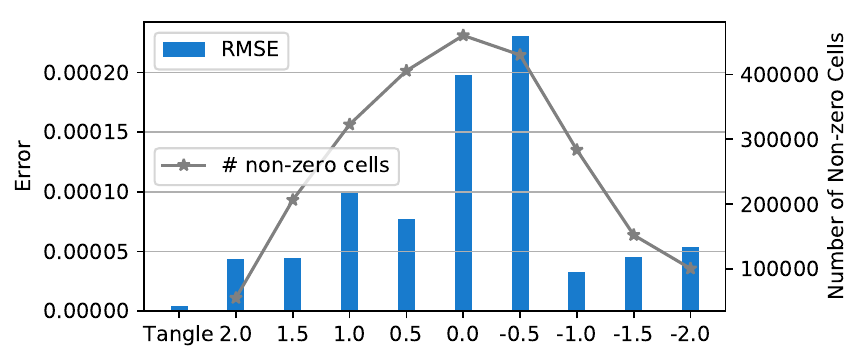}
  \caption{%
    LCP field average error using our approach on the tangle dataset (the first column) and for different iso-values on the ethanediol dataset (the second to the 10th columns). We also use a line chart to plot the number of non-zero probability cells to compare the trend.
  } 
  \label{fig:acc}
\end{figure}

We show both the quantitative and qualitative results of the accuracy of our approach in this section. 
The quantitative results (root mean squared error) are shown by the blue bars in \cref{fig:acc}.
For the tangle dataset, the root mean squared error between the ground truth (dense reconstruction method) and our approach is $4.6 \times 10 ^{-6}$. For the ethanediol dataset, the root mean squared error ranges from $3.2 \times 10^{-5}$ to $2.3 \times 10^{-4}$. Given that the probabilities are in the range $[0,1]$, the average error of our approach is minimal. Errors are introduced by the threshold $\alpha$ chosen to classify a cell to have zero probability. Therefore, when the iso-surface is complex indicated by more non-zero cells, our accuracy tends to drop slightly. This is confirmed by the results in \cref{fig:acc} that the RMSE is higher when the number of non-zero cells is relatively large (iso-values equal 1.0, 0.5, 0.0, and $-0.5$). However, the errors may not strictly follow this trend, for there are other errors introduced by local SGP approximation and L-BFGS-B not finding the global minima.

\begin{figure}[htb]% specify a combination of t, b, p, or h for the top, bottom, on its own page, or here
  \centering % avoid the use of \begin{center}...\end{center} and use \centering instead (more compact)
  \includegraphics[width=\columnwidth]{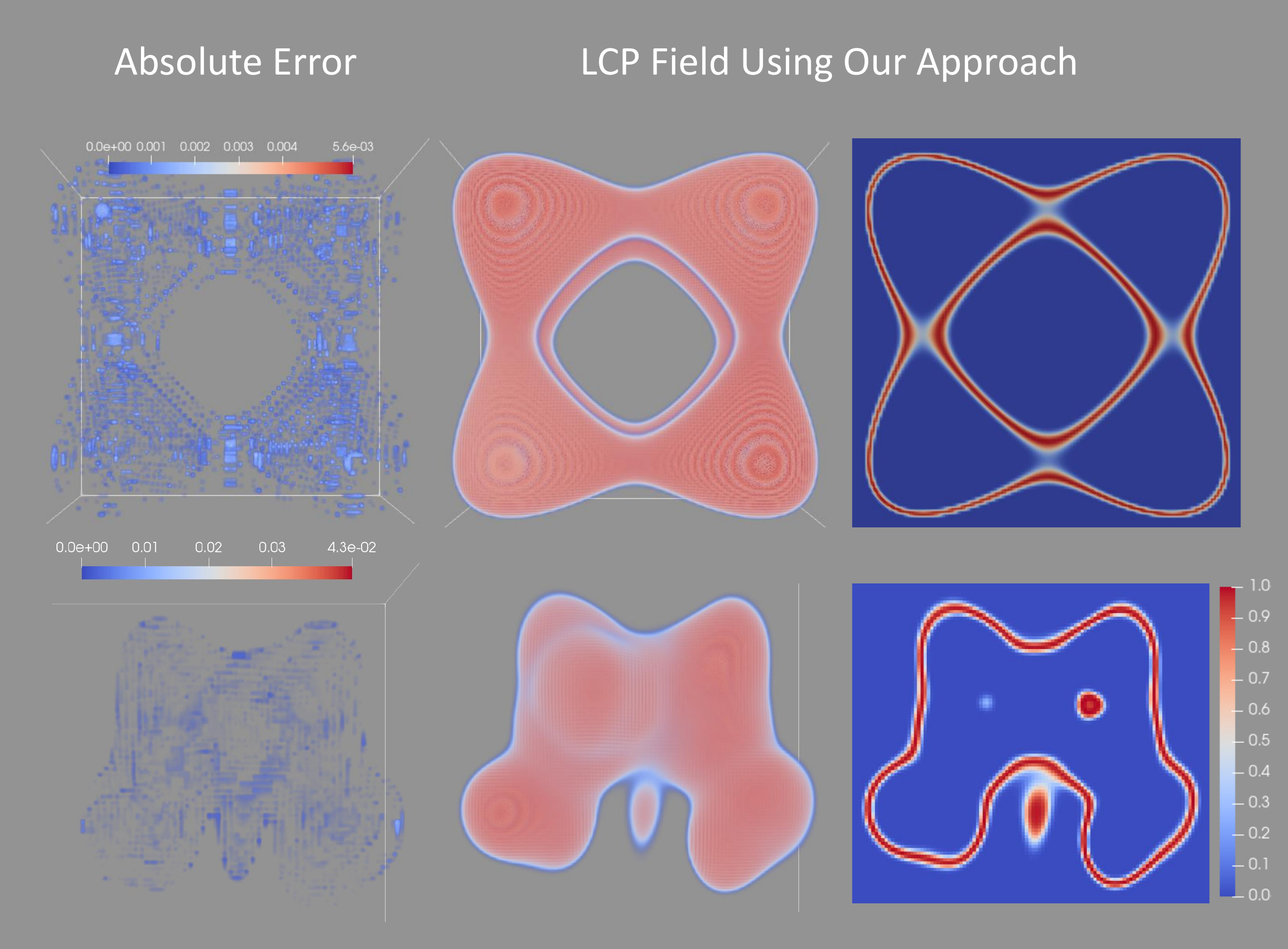}
  \caption{%
    The two figures on the left show the absolute error between our approach and the baseline method, the two figures in the middle show the calculated 3D LCP field using our approach, and the two figures on the right show the 2D slides of the LCP field. The color mapping of the absolute error is shown by color bars beside each figure, while the color mapping of the probability is shown by the color bar on the right.
    It can be seen most errors occur at the edges where the probability fades to zero.
  } 
  \label{fig:diff}
\end{figure}

In terms of qualitative results shown in \cref{fig:diff}, we calculate the absolute error between the LCP field calculated using our method and the LCP field from dense reconstruction (ground truth). The volume rendering of this difference field is shown in the left column.
We observe that most of the errors occur at the boundaries where the probability fades to zero. This is also because the threshold we choose will inevitably classify some nodes with a very small probability to zero probability. These nodes with very small probabilities usually occur on the boundary of the surface.

In summary, the errors introduced by our acceleration method are generally very small for different iso-values and they are controllable by a hyperparameter $\alpha$, the effect of changing which will be discussed in \cref{sect:hyper_parameters}. We will also discuss the consequences of errors in detail in \cref{sect:discussion}.

\subsection{Hyperparameters}
\label{sect:hyper_parameters}

There are two important hyperparameters $\alpha$ and $\beta$ to pick in our approach. The hyperparameter $\alpha$ is the threshold to classify a node as zero probability, which means when the estimated probability value is smaller than $\alpha$, the node will not be subdivided anymore. The hyperparameter $\beta$ is the threshold for local SGP, which should be chosen large enough to make local SGP similar to the original SGP model. We perform the experiments to show the effects of these two hyperparameters in this section and make suggestions for their values. 

\begin{figure}[htb]% specify a combination of t, b, p, or h for the top, bottom, on its own page, or here
  \centering % avoid the use of \begin{center}...\end{center} and use \centering instead (more compact)
  \includegraphics[width=\columnwidth]{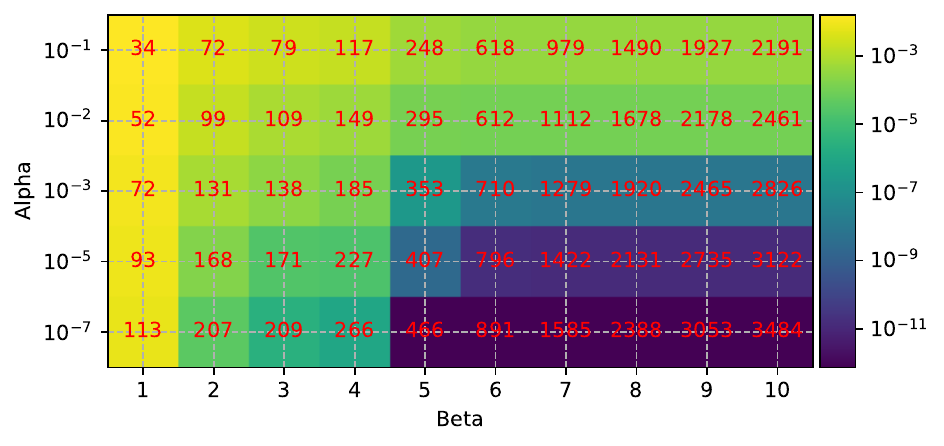}
  \caption{%
    Grid search results of the hyperparameter $\alpha$ and $\beta$. Color in the figure shows the mean squared error (MSE) of the LCP field calculated using the corresponding hyperparameters. We use a log scale to differentiate tiny MSE differences. The numbers inside show the computation time under each hyperparameter setting.
  } 
  \label{fig:hyper}
\end{figure}

We perform a grid search to tune the hyperparameters. To show the tuning results, we visualize a heatmap of the mean squared error under different hyperparameter settings in \cref{fig:hyper}. The dataset used in the tuning is Ethanediol and we use iso-value $-1.5$ for the grid search. However, we will explain how this result can be generalized to other datasets and iso-values.

We first examine the search results for $\beta$. We compare $\beta$ with $\frac{\norm{\bm{x}_1-\bm{x}_2}}{\ell}$, where $\norm{\bm{x}_1-\bm{x}_2}$ is the euclidean distance between two positions and $\ell$ is a GP kernel hyperparameter called length scale that is already tuned in the model fitting process. In other words, we want to only consider the inducing points in a radius less than $\beta\ell$ to build the local SGP model. Because the length scale hyperparameter $\ell$ can be found in all kinds of GP kernels and $\frac{\norm{\bm{x}_1-\bm{x}_2}}{\ell}$ can be seen as a normalized distance for this specific model, the suitable hyperparameter $\beta$ found in this experiment is generalizable to other datasets and models. The x-axis in \cref{fig:hyper} denotes the change of $\beta$ from 1 to 10. We can observe that the MSE drops when we increase $\beta$ from 1 to 6, but no longer drops when $\beta$ is larger than 6, which indicates 6 is a sweet point for computation efficiency and accuracy.

Next, we discuss the effect of changing $\alpha$, which is the threshold to classify a node to have zero level-crossing probability. 
In the experiments, we tested $\alpha = 10^{-1}, 10^{-2}, 10^{-3}, 10^{-5}, 10^{-7}$.
In the y-direction of \cref{fig:hyper}, we can see that the MSE drops when we decrease $\alpha$ from $10^{-1}$ to $10^{-7}$. However, although the MSE continues to drop, the computation cost increases because we consider more cells in the calculation that have a tiny probability. The choice of $\alpha$ depends on the precision we want to achieve in the LCP calculation, which is also a trade-off between precision and computation cost. In our experiments, we set $\alpha = 10^{-3}$, which classifies the probability that is less than $10^{-3}$ to be zero (MSE is approximately $10^{-6}$). The choice of $\alpha$ only depends on the precision of LCP we want to achieve and does not depend on the iso-value and the dataset.

\subsection{Scalabilty}
\label{sect:scalability}

The scalability is evaluated from two perspectives. The first is our method's scalability to the reconstruction resolution and the second is our method's scalability to the SGP model size. We choose the target resolutions of 64, 128, 256, 512, and 1024 to test our method's scalability on the ethanediol dataset and report the calculation time and accuracy under these resolutions. The results are reported in \cref{tab:res}.
\begin{table}[ht]
  \caption{
  This table shows the calculation time and accuracy for different resolutions on the ethanediol dataset with iso-value $-1.5$. Numbers with stars are theoretical time since the experiment under that condition takes too long to finish. We also cannot calculate the errors in those cases because the dense results are not available.
  }
  \label{tab:res}
\begin{tabular*}{\columnwidth}{@{\extracolsep{\fill}}c c c c c c c}
  \hline
  Res.   & $64$  & $128$ & $256$ & $512$ & $1024$  \\
  \hline 
  Our    & 46.7 s & 4.2 m & 20.3 m & 2.17 h & 14.1 h \\
  Dense  & 156.5 s & 20.9 m  & 164.9 m & 22.0 h * & 7.3 d * \\
  RMSE ($ 10^{-5}$)   & 0.7 & 4.5 & 4.6 & - & - \\
  \hline 
\end{tabular*}
\end{table}
We can see clearly in the table that our computation time is significantly shorter than the baseline. The percentage of reduction increases with the increase of the resolution. The errors are also constantly low for different resolutions. For the mid-to-high-resolution cases ($512$ and $1024$), the dense reconstruction cannot be finished in practical time. Therefore, we only show the theoretical computation times for dense reconstruction. 
Although we cannot directly compare the reconstruction results between our method and the baseline to get the errors for these resolutions. \clrb{The errors of other resolutions indicate our accuracy is constantly high and our approach makes the calculation of LCP practical in high resolutions.}

In terms of the SGP model size, we train three different models on the ethanediol dataset with $100$, $250$, and $500$ inducing points. Similarly, we report the LCP calculation time and the accuracy in \cref{tab:model_size}. 
\begin{table}[ht]
  \caption{
  The computation time and accuracy comparison under different model sizes (different number of inducing points). We report the time in seconds and the percentage of our approach compared to the baseline. The accuracy is reported as the root mean squared error. 
  }
  \label{tab:model_size}
\begin{tabular*}{\columnwidth}{@{\extracolsep{\fill}}c c c c}
  \hline
\# Inducing Points &$100$  & $250$  & $500$  \\
  \hline 
  Our Time (s) & 72.0 (20.7\%) & 109 (18.3\%) & 249 (20.2\%) \\
  Dense Time (s) & 347.6 & 597 & 1232 \\
 RMSE  & $0.6\times10{-4}$ & $2.4\times10{-4}$ & $0.5\times10{-4}$ \\
  \hline 
\end{tabular*}
\end{table}
We can see that the computation time for our approach is consistently less than the computation time of the dense reconstruction. The percentage of time our method takes compared to the dense reconstruction is roughly $20\%$. The calculation error is consistently low across different model sizes. \clrb{The errors are not necessarily monotonically decreasing when increasing the model size because they are calculated between the reconstructed LCP field from our approach and the dense method but not between reconstruction and the original data before applying the SGPR model.}

In summary, our approach is scalable to higher target resolution and larger SGPR models with more inducing points with the computation time significantly lower than the dense reconstruction and the error being very small.

\subsection{Adaptive Query Pattern}
\label{sect:query_pattern}
\begin{figure}[htb]% specify a combination of t, b, p, or h for the top, bottom, on its own page, or here
  \centering % avoid the use of \begin{center}...\end{center} and use \centering instead (more compact)
  \includegraphics[width=\columnwidth]{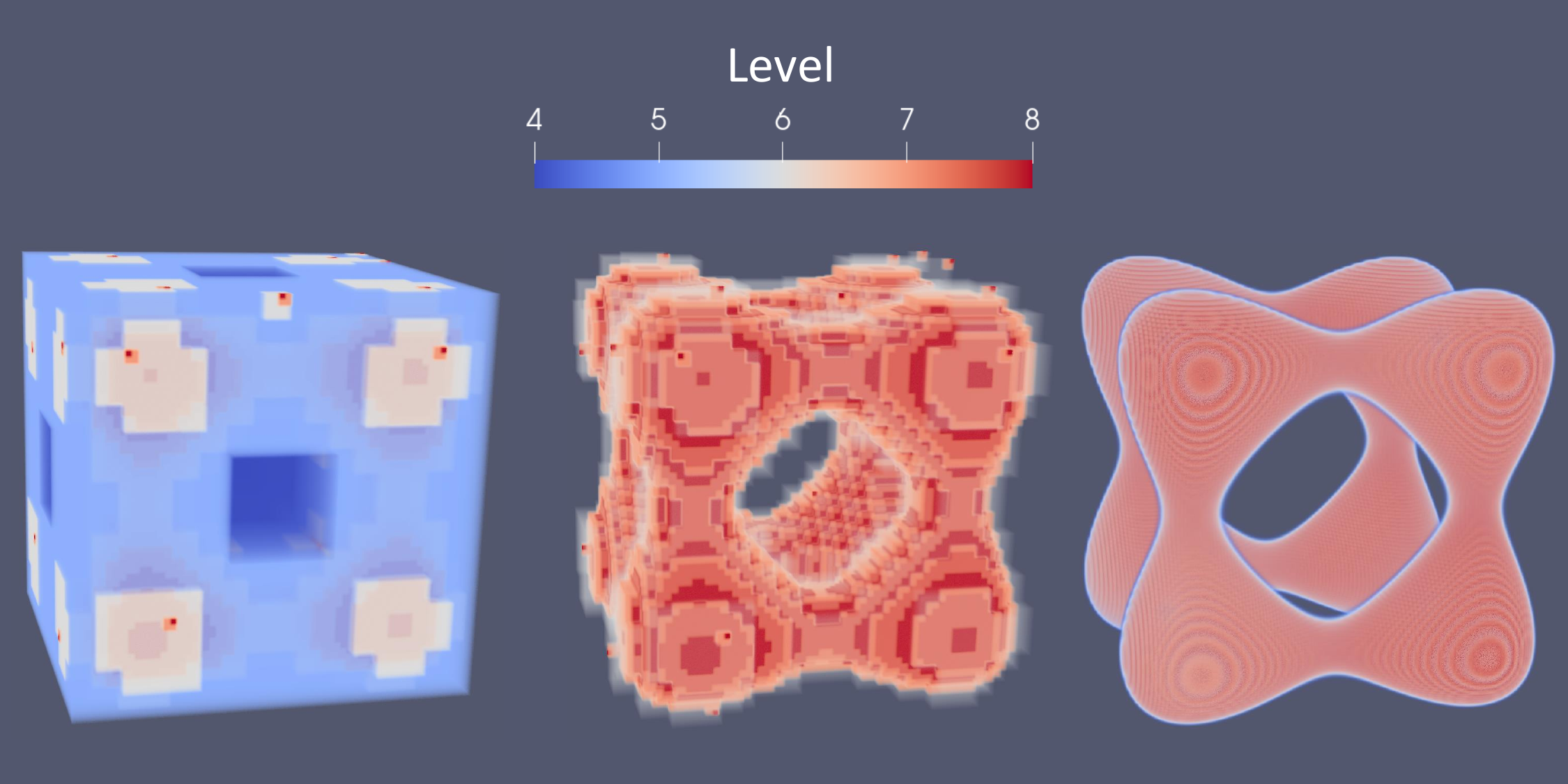}
  \caption{%
    The left and the middle figures show the volume rendering result for the octree level. Red indicates a deeper level and blue indicates a more shallow level. We change the opacify in the transfer function to highlight different levels. We show the calculated LCP field on the right for a comparison. 
  } 
  \label{fig:pattern}
\end{figure}

As the last part of the evaluation, we visualize the adaptive query pattern using volume rendering to show that our method indeed focuses its calculation on the cells where the level-crossing probability is not zero.
After the execution of \cref{alg:subdivision}, we are left with an octree showing the query pattern. This tree has a maximum depth related to our chosen target resolution. However, only regions that our algorithm estimates to have non-zero probability will be subdivided to the maximum depth, while other regions will be skipped at some level of this tree. Visualizing the levels of the leaf nodes (where we stop subdivision) will demonstrate the query pattern of our approach. We show the volume rendering result in \cref{fig:pattern}. 

The middle image in \cref{fig:pattern} highlights the deeper levels of the octree. We can see that most deep-level nodes are distributed around the places where level-crossing probabilities are high. In the left image, we show the shallow levels of the octree, which corresponds to the skipped node in the adaptive query. These skipped nodes are distributed at the empty spaces where the LCP is indeed zero.
These observations confirm the rationale of our approach that we skip the regions with zero level-crossing probability and focus the computation on the non-zero-probability regions.

%-------------------------------------------------------------------------
\section{Example Usage in GP Uncertainty Analysis}
\label{sect:application}

In this section, we demonstrate how we perform uncertainty analysis using our method on the data represented by a sparse Gaussian process model. We want to emphasize that even though calculating LCP to perform uncertainty analysis is possible without our approach, the long computation time for GP inference and Monte Carlo sampling makes it prohibitive in practice (especially for the high resolution as shown in \cref{sect:scalability}). 

\begin{figure}[htb]% specify a combination of t, b, p, or h for the top, bottom, on its own page, or here
  \centering % avoid the use of \begin{center}...\end{center} and use \centering instead (more compact)
  \includegraphics[width=\columnwidth]{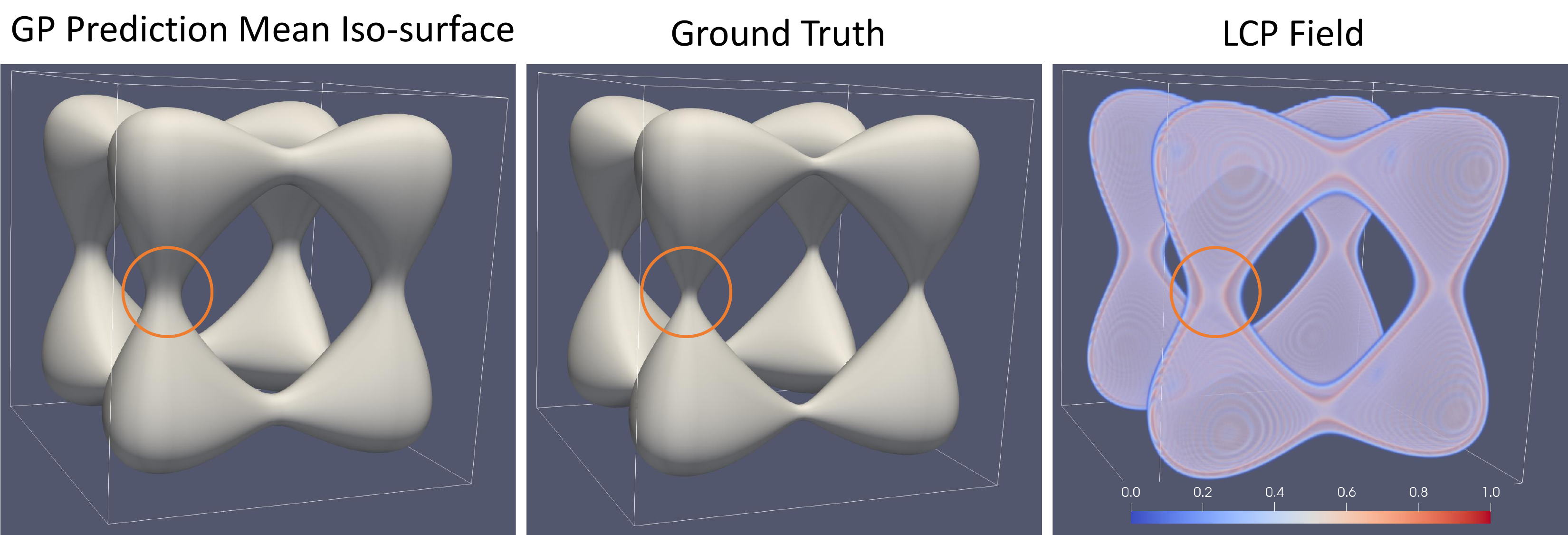}
  \caption{%
    We visualize the iso-surface extracted from the GP predicted mean field on the left, from the ground truth data in the middle, and the calculated LCP field on the right. The differences in the ``bridge'' structure are highlighted by the orange circles.
  } 
  \label{fig:uq_1}
\end{figure}

For the Tangle dataset, we are interested in the uncertainty of the bridge structure as highlighted by the orange circle in \cref{fig:uq_1}.
We extract the iso-surfaces from the GP-predicted mean field (left) and compare it to the ground truth iso-surface (middle). We can see apparent differences in the ``bridge'' structures highlighted by the circle in the figure. The predicted ``bridge'' is much wider than the ground truth ``bridge''. This uncertainty could be misleading since we no longer have access to the original ground truth data if the GP model is used for data reduction purposes. Luckily, the information for uncertainty is also available from the GP model. We visualize the LCP field using our method and visualize it with volume rendering (the right image in \cref{fig:uq_1}). The transfer function is edited to highlight the medium probabilities (0.25 to 0.5). We can see there is a wider range of medium probability in the region of the ``bridge'' structure, indicating that the GP model is not certain about its prediction in this region. Having this knowledge, we can adjust our trust in different regions of the GP-predicted field and make more reliable conclusions.

\begin{figure}[htb]% specify a combination of t, b, p, or h for the top, bottom, on its own page, or here
  \centering % avoid the use of \begin{center}...\end{center} and use \centering instead (more compact)
  \includegraphics[width=\columnwidth]{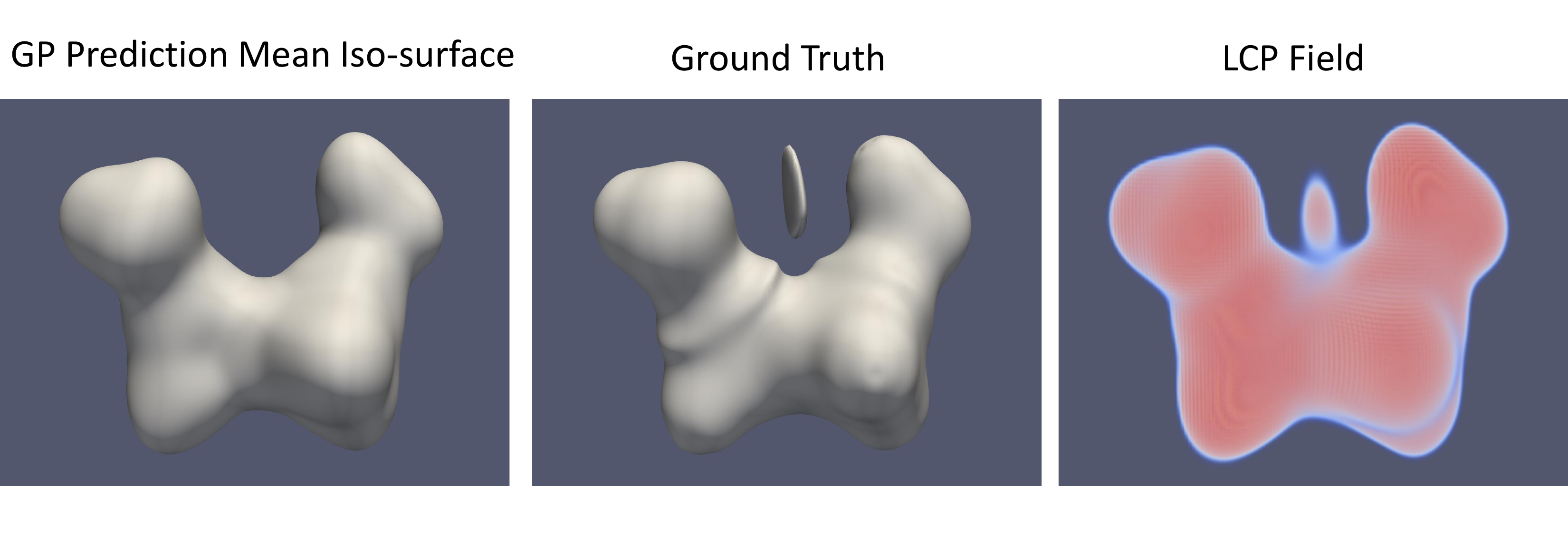}
  \caption{%
    The visualization of the iso-surface extracted from the GP predicted mean field on the left, from the ground truth data in the middle, and the calculated LCP field on the right. Notice the missed disk-shape structure in the GP predicted mean iso-surface.
  } 
  \label{fig:uq_2}
\end{figure}

We perform a similar analysis on the ethanediol dataset which is obtained from a real simulation. We visualize the iso-surface with value $-1.5$ and the results are shown in \cref{fig:uq_2}.
We can easily identify the difference between the GP reconstruction and the original data. The reconstruction does not have a disk-shaped structure in the middle. This could lead to dangerous conclusions when using the reduced GP model for analysis. Similarly, we visualize the LCP field for this reconstruction and find that the disk-shaped structure is possible to exist in the reconstruction field. However, only visualizing the mean field will miss this feature. 

In summary, visualizing the level-crossing probability on the dataset represented by a GP model can reveal the GP's uncertainty to the fitting of the original data. The technique presented in this paper reduces the computation cost for the LCP calculation and makes it practical to use.
%-------------------------------------------------------------------------

\section{Discussion and Future Works}
\label{sect:discussion}

We first discuss some possible issues and shortcomings in this section, which includes different ways for adaptive queries, local SGP approximation, situations when our method could fail, and the adaptation to other types of uncertainty models.

% kd-tree
In this paper, we present the adaptive query pattern using an octree data structure. However, this may not be the optimal choice. Using a K-D tree could skip a larger block because instead of splitting simultaneously in three dimensions, a K-D tree splits the dimensions in turn. Skipping a larger block with zero probability can reduce the overhead in our approach. In practice, a K-D tree version of our algorithm can be beneficial.

%local SGP approximation
We proposed a local SGP approximation to reduce the inference time in the probability estimation, however, by only taking the example of RBF kernel in \cref{sect:local_sgp}. Theoretically, the definition of the local SGP can be generalized to any kind of kernel which is an isotropic and monotonic function of the distance. However, for periodic kernel functions, the local SGP cannot be built using a simple distance threshold. We need to calculate the covariance for every inducing point and compare this covariance to a threshold. It is also not easy to find a single local SGP for a node region in the octree, because the inducing points with large covariance may change a lot in the region. In summary, it requires more careful design to build local SGPs for periodic kernel functions which are not fully discussed in this paper.

%fail cases
There are some special edge cases where our approach may work poorly in time or accuracy. Because our method uses adaptive sampling to find non-zero probability cells to reduce the total computation, if an LCP field has non-zero probability everywhere in the volume, our approach can be even slower than the dense reconstruction due to the overhead. Luckily, this rarely happens in real-world datasets. 
Another potential issue lies in the process of finding the minimum probability in a region using L-BFGS-B. L-BFGS-B may not converge to a global minimum when the target function $F_M(\bm{x})$ is not convex. 
In that case, the real probability can be larger than this upper bound and we may misclassify a region to have zero probability. In the resulting field, this could lead to large errors because we assign zero probabilities to a region that should not be zero. 

% other types of uncertainty other than Gaussian
\clrb{Finally, since our acceleration approach is built based on the assumption the original data is represented by a Gaussian process model, the uncertainty of the data can only be modeled using Gaussian distributions. Other types of uncertainty that are also common in scientific data are not discussed in the evaluation because of this restriction. Furthermore, more sophisticated GP models used on complex scientific datasets, like hierarchical SGPR are also not evaluated in this paper, we plan to investigate them and extend the applicability of this work in the future.}

With the shortcomings of our approach discussed, our future works aim to solve these problems. The probability approximation method proposed in the previous paper \cite{pothkow2013approximate} can be used to derive a tighter upper bound for regional level-crossing probability. A tighter upper bound can lead to more efficient zero-probability cell skipping and reduce the total calculation time. 
K-D trees can be utilized in future work to further reduce the computation time. Moreover, we are in search of other approaches instead of L-BFGS-B to find the global minimum probability in the region so that we can guarantee the upper bound estimation theoretically.

\section{Conclusion}

We present an efficient level-crossing probability field calculation method for GP-represented data through the adaptive query. To facilitate fast regional level-crossing probability estimation, we derive an upper bound for the regional level-crossing probability and use a local SGP model and L-BFGS-B algorithm to calculate this upper bound efficiently.
We demonstrate that our method takes a much shorter time to calculate the LCP field compared to the dense reconstruction with high accuracy.

\acknowledgments{
This work is supported in part by the National Science Foundation Division of Information and Intelligent Systems-1955764, the National Science Foundation Office of Advanced Cyberinfrastructure-2112606, U.S. Department of Energy Los Alamos National Laboratory contract 47145, and UT-Battelle LLC contract 4000159447 program manager Margaret Lentz.}

\bibliographystyle{abbrv-doi}

\bibliography{template}
\end{document}